\documentclass{article}

\usepackage[preprint,nonatbib]{neurips_2021}

\usepackage[utf8]{inputenc} 
\usepackage[T1]{fontenc}    
\usepackage{hyperref}       
\usepackage{url}            
\usepackage{booktabs}       
\usepackage{amsfonts}       
\usepackage{nicefrac}       
\usepackage{microtype}      

\usepackage{booktabs} 
\usepackage{subcaption}
\usepackage{amsmath}
\usepackage{mathtools}
\usepackage[toc,page]{appendix}
\usepackage{enumitem}
\usepackage{graphics}
\usepackage{graphicx}
\usepackage{algorithm}
\usepackage{algorithmicx}
\usepackage[noend]{algpseudocode}
\usepackage{enumitem}
\algnewcommand{\LineComment}[1]{\State \(//\) #1}
\DeclareMathOperator*{\mode}{mode}

\usepackage[export]{adjustbox}

\title{Continual Competitive Memory: A Neural System for Online Task-Free Lifelong Learning}

\author{%
  Alexander Ororbia \\
  Rochester Institute of Technology \\
  Rochester, NY, USA \\
  \texttt{ago@cs.rit.edu}
}

\begin{document}
\setlength{\abovedisplayskip}{0.065cm}
\setlength{\belowdisplayskip}{0pt}

\maketitle

\begin{abstract}
\label{sec:abstract}
In this article, we propose a novel form of unsupervised learning, continual competitive memory (CCM), as well as a computational framework to unify related neural models that operate under the principles of competition. The resulting neural system is shown to offer an effective approach for combating catastrophic forgetting in online continual classification problems. We demonstrate that the proposed CCM system not only outperforms other competitive learning neural models but also yields performance that is competitive with several modern, state-of-the-art lifelong learning approaches on benchmarks such as Split MNIST and Split NotMNIST. CCM yields a promising path forward for acquiring representations that are robust to interference from data streams, especially when the task is unknown to the model and must be inferred without external guidance.
\end{abstract}	


\section{Introduction}
\label{sec:intro}

Lifelong machine learning, otherwise known as continual or never-ending learning \cite{thrun1995lifelong,mitchell2018never}, stands as one of the greatest challenges facing statistical learning research, especially for models based on artificial neural networks (ANNs). In this problem context, the goal of an agent is much harder than just learning one single prediction task from one single dataset; it must, instead, learn multiple, different prediction tasks \emph{across} several datasets, much as human agents do. In order to do so, the agent must be able to aggregate and transfer its knowledge as new pattern vectors from new problems are encountered.
ANNs particularly struggle to continually learn due to the well-known fact that they tend to \emph{catastrophically forget} \cite{mccloskey_catastrophic_1989,ratcliff_connectionist_1990,french1999catastrophic,lewandowsky1994relation}, or rather, they completely erase the knowledge acquired from earlier encountered tasks when processing samples from new tasks.
In essence, the problem of catastrophic forgetting (or interference) becomes prominent when the data being presented to an ANN is no longer independently and identically distributed, a fundamental assumption that drives most modern-day machine learning machinery, particularly systems based on deep ANNs.
This kind of forgetting imposes practical issues and impedes the use of ANNs as potential models of mind. 

While a great deal of research, including both classical and modern efforts, have attempted to address the problem of catastrophic forgetting (or interference), introducing regularization frameworks \cite{kirkpatrick2017overcoming}, new mechanisms that allow networks to grow/expand or prune their internal synaptic structures \cite{rusu2016progressive,parisi2017lifelong}, or forms of pattern rehearsal \cite{robins1993catastrophic,robins1995catastrophic,rebuffi2017icarl}, most if not all of these efforts make the use of external human guidance in the form of task-descriptors, or special vectors that indicate the differences between task datasets. Task descriptors themselves are not necessarily neuro-cognitively unrealistic, given that when asked to solve high-level problems, humans are often given some indicator of the task to be completed (either through audio-visual cues, environmental context, or verbal communication). However, for the relatively low-level pattern recognition problems that deep neural networks are often evaluated on, e.g., object recognition, image categorization, generative modeling of low-level features like pixels, it seems rather unlikely that systems of real neurons would be provided, from some external arbiter, such explicit information in order to identify the differences between such tasks.

In this work, we tackle the far greater challenge imposed by continual learning -- learning \emph{without} external task descriptors at \textbf{both} training and test time phases. Some of the few, more recent efforts that have attempted to tackle such a problem \cite{aljundi2018selfless} have labeled this scenario as ``selfless'' or task-free (sequential) learning, motivating the need for agent systems that learn without task boundary information. Crafting continually learning agents that can would have strong implications for applications ranging from space exploration robotics to autonomous vehicles. 

Under the problem context above, our contributions in this article are as follows:
\begin{itemize}[noitemsep,nolistsep]
    \item We craft a simple modeling framework -- the neural competitive learning (NCL) framework -- for unifying neural models that learn through competitive (Hebbian) learning and examine  several important ones in the online streaming setting, such as classical adaptive resonance theory. 
    \item We propose a novel, unsupervised learning system, i.e., \emph{continual competitive memory}, that builds on the key ideas of competitive learning systems and is generalized for the purpose of continually learning from data streams without task descriptors.
    \item We measure the amount of forgetting in our CCM model and learning process using several metrics that test knowledge retention as well as class conformity across several low-level pattern classification tasks comparing to important NCL baselines as well as a backprop-based neural model.
\end{itemize}
Our experimental results indicate that the proposed CCM system not only outperforms the baselines tested and examined but also yields, when it is allowed to drive a feedforward neural predictor, competitive performance on two key benchmarks, Split MNIST and Split NotMNIST. Our general NCL framework as well as our proposed CCM neural model offer a foundation upon which future research in both growing and fixed-capacity neural systems for task-free continual learning may be built on top of.

\section{The Neural Competitive Learning Framework}
\label{sec:ncl}

We start by first defining the variant of the continual learning problem this work aims to tackle. We then will present a general framework that serves as a modern-day generalization of the classical principles put forth in \cite{rumelhart1985feature}, which we call the neural competitive learning (NCL) framework, and unifies the types of neural models that we will design for the problem setting. The NCL framework will also be used to distinguish the key components required by adaptive artificial neural systems that could be classified as ``competitive''.

\subsection{Desiderata for Streaming Lifelong Learning Systems}
\label{sec:lml_criteria}
The form of continual learning that is of interest to this study, and simultaneously one of the problem's hardest forms, satisfies the following criteria:
\begin{enumerate}[noitemsep,nolistsep]
    \item \textbf{Data is processed sequentially online:} the agent receives samples, one at a time (or, at best, in small mini-batches), from a potentially infinite stream of data. Once a datum has been seen, it will not be stored and is not directly available ever again.
    \item \textbf{Task descriptors are not provided:} the agent is not provided with any supervised signal from the experimenter as to what task it is operating on, not even an encoded integer identifier (as is common in many continual learning setups). This condition holds at both training and testing times.
    \item \textbf{Raw sensory data is not be stored:} the agent should not physically store data patterns (as in the case of replay buffers common in rehearsal-based approaches) or, at worst, store a temporary, negligible quantity at any instant (to simulate working memory). This feasibility requirement ensures that the agent's memory footprint does not grow uncontrollably with respect to the size of the stream (which could be infinite).
    \item \textbf{Patterns from the stream might or not come with labels:} the agent should not rely on labels to self-organize as there is no guarantee that the stream will provide labels for each and every single data point it is to produce. Nevertheless, even if the agent is to use labels, the agent is still expected to learn and operate even when parts of the stream are unsuperivsed.
\end{enumerate}
The NCL models that we examine, develop, or design in the rest of the paper satisfy all four of the above criterion.
Inspired by the setup discussed in \cite{van2019three}, our desired set of criterion above means that NCL systems handle the class-incremental learning (Class-IL) scenario, where an agent must solve each task seen so far as well as infer which task it is being confronted with.

\subsection{Preliminaries}
\label{sec:prelim}

\paragraph{Notation: }\label{sec:notation}
In this paper, $\otimes$ indicates a Hadamard product while $\cdot$ indicates a matrix/vector multiplication (or dot product if the two objects it is applied to are vectors of the exact same shape) and $(\mathbf{v})^T$ denotes the transpose.
$||\mathbf{v}||_p$ represents a $p$-norm (also used to represent a distance function, as in the case when the difference between two vectors is provided as the argument) -- $p = 1$ chooses the Manhattan (L1) distance and $p = 2$ selects the Euclidean (L2) distance.
$|\Theta|$ denotes the cardinality of the set $\Theta$. $\text{r}(\mathbf{v})$ and $\text{r}(\mathbf{M})$ return the number of rows in vector $\mathbf{v}$ and in matrix $\mathbf{M}$, respectively, while $\text{c}(\mathbf{v})$ and $\text{c}(\mathbf{M})$ return the number of columns in $\mathbf{v}$ and $\mathbf{M}$, respectively.

$\mathbf{M}[:,i]$ is the slice operator, meant to extract the $i$th column vector of matrix $\mathbf{M}$ -- note that $\mathbf{M}[i,:]$ means extract the $i$th row. $\mathbf{M}[j,i]$ means extract the scalar element/value in the position $(j,i)$ in $\mathbf{M}$.
In this paper, we will sometimes refer to a higher-order tensor (usually three-dimensional/3D) and retrieving the $i$th 2D slice is denoted as $\mathbf{M}[i,:,:]$.

$\mathbf{z} = \mbox{Enc}(w, d_z)$, the ``one-hot encoding'' function, means convert a list of integers $w$ into a binary vector $\mathbf{z}$, where each value in $\mathbf{z}$ is one at each integer/index inside of $w$ and zero elsewhere. Note that $\arg\min_i \mathbf{v}$ returns the index $i$ of $\mathbf{v}$ with the minimum value and, conversely, $\arg\max_i \mathbf{v}$ returns the index of $\mathbf{v}$ with maximum value.

We also introduce the following useful operators/macros. Given column vector $\mathbf{v} \in \mathcal{R}^{r \times 1}$ and row vector $\mathbf{u} \in \mathcal{R}^{1 \times c}$, we define the following:
\begin{itemize}
    \item $\mathbf{o} = \text{BC}_{c}(\mathbf{v}) = \mathbf{v} \cdot \mathbf{1}_c$, where $\mathbf{1}_c = \{1\}^{1 \times c}$. The result is a matrix $\mathbf{o} \in \mathcal{R}^{r \times c}$ with $\mathbf{v}$ copied into each column ($\mathbf{v}$ is replicated column-wise $c$ times),
     \item $\mathbf{o} = \text{BR}_{r}(\mathbf{u}) = \mathbf{1}_r \cdot \mathbf{u}$, where $\mathbf{1}_r = \{1\}^{r \times 1}$. The result is a matrix $\mathbf{o} \in \mathcal{R}^{r \times c}$ with $\mathbf{u}$ copied into each row ($\mathbf{u}$ is replicated row-wise $r$ times).
\end{itemize} 
Note that the above two operators are typically implemented internally in some programming languages as broadcasting/replication.

\paragraph{Problem Setting: }\label{sec:problem}
Any of the neural systems we will investigate in this study are to be applied to streams of pattern vectors following the form $\{ (\mathbf{x}_1,\mathbf{y}_1), (\mathbf{x}_2,\mathbf{y}_2),...,(\mathbf{x}_N,,\mathbf{y}_N) \}$ where any single input vector $\mathbf{x}_j$ contains $d_x$ dimensions, i.e., $\mathbf{x}_j \in \mathcal{R}^{d_x \times 1}$, and any single label vector contains $d_y$ dimensions, i.e., $\mathbf{y}_j \in \{0,1\}^{d_y \times 1}$ (typically $\mathbf{y}_j$ is the one-hot encoding of a class index $y_j$). While $N$ does indicate that the data stream is finite and terminates after $N$ steps, it is more likely in real-world problem settings that the stream would be infinite, i.e., $N = \infty$. Each data point could come from one of $T$ possible tasks/datasets (in the experiments presented later, we will investigate benchmarks with $T=5$ tasks).

Note that the neural models we consider and design in this work are unsupervised in nature, i.e., they do not require or use class labels to conduct learning and/or inference (whereas models based on learning vector quantization \cite{kohonen1995learning} do require labels), easily satisfying the fifth criterion of Section \ref{sec:lml_criteria}. Nonetheless, later we will introduce simple extensions that will allow an NCL model or our proposed CCM to conduct classification given that the benchmarks we evaluated on contain ground truth labels. Such extensions will allow us to evaluate model quality and generalization in greater detail.

\subsection{General Criterion and Organization}
\label{sec:general_model}

Our framework -- the Neural Competitive Learning (NCL) framework -- breaks down competitive neural systems into two key components: 1) a best matching unit function, and 2) a synaptic weight adjustment scheme. These are the two essential pieces that must be designed to create a minimally functional NCL model.

Our motivation behind investigating and generalizing competitive learning to the continual learning problem space is ultimately to reduce excessive cross-talk \cite{willshaw2014holography} (which refers to the information capacity of an ANN) -- catastrophic interference is effectively cross-talk with a ``vengeance'' \cite{french1991using}. Classical work \cite{french1991using} argued that activation overlap -- the average shared activation value over all neural units in a given layer -- can create the cross-talk that either places the ANN beyond its information capacity (thus resulting in the destruction of old knowledge) or results in the situation where learning even a single new input can disrupt previously acquired information. One path for tackling the activation overlap problem would be to bias the ANN to acquire semi-distributed representations -- a ``sweet-spot'' between fully distributed representations that generalize well (but lead to catastrophic forgetting) and local representations that are immune to forgetting (but do not generalize well).
Our intuition is that, by using representations that are driven by neuronal units that fight for the right to activate in response to particular input stimuli, naturally emergent semi-distributed representations will emerge from the complex system. These representations will be far less likely to collide (provided that the right mechanisms are built into the memory system) yet still generalize well (since many possible input configurations will still match to a given memory unit).

Furthermore, competition across units creates sparse activity patterns. Since sparsity has been shown to prevent forgetting in classical systems such as sparse distributed memory \cite{kanerva1988sparse} (before saturation), we argue that a carefully designed neural memory system that utilizes competition among its units can effectively be used to tackle the difficult problem of task-free continual learning described earlier.

\begin{algorithm}[!t]
\caption{The algorithm for selecting the $K$ best matching units.}
\label{algo:competition}
\begin{algorithmic}
   \State {\bfseries Input:} Pre-activation vector $\mathbf{h}$, $K$ BMU desired, winner\_type string flag
   \Function{FindBMU}{$\mathbf{h}$, $K$, winner\_type} \Comment Run $K$-winners selection process
        \State $w = \emptyset$,  $h_{min} = \min(\mathbf{h})$, $h_{max} = \max(\mathbf{h})$
        \Comment{Initialize statistics}
        \For{$k = 1$ to $K$}
        \If{winner\_type \textbf{is} ``max''} \Comment winner\_type is unit w/ maximum value
            \State $i = \arg\max_i \mathbf{h}$, \; $\mathbf{h}[1,i] = h_{max}$
        \Else \Comment winner\_type is  unit w/ minimum value (``min'')
            \State $i = \arg\min_i \mathbf{h}$, \; $\mathbf{h}[1,i] = h_{min}$
        \EndIf
        \State $w \leftarrow w \cup \{i\}$ \Comment Record $k$th index of $k$th winning neuron
        \EndFor
        \State \textbf{Return} $w$ \Comment Note, in this study, we treat $\{i\} = i$ 
    \EndFunction
\end{algorithmic}
\end{algorithm}

\paragraph{Best Matching Unit (BMU) Activation Function: }
The most fundamental and necessary component for any neural system to be considered competitive is the design of a function for selecting a one or more units out of a pool of neurons, conditioned on sensory input or external activation. In basic competitive neural models, this is known as the \emph{best matching unit} (BMU) and often refers to selecting one single neuron\footnote{We will, throughout the paper, also refer to these as ``prototypes'' or memory units/slots.}, which satisfies a particular criterion/constraint, out of many. However, a good deal of work in competitive learning often involves computing more than just one single neuron (at least the second BMU) (e.g., growing neural gas \cite{fritzke1995growing}) and our framework accounts for this by allowing more than one unit to potentially win the competition. Selecting a winning neuron can either be done by selecting the maximum, i.e., the $\max( )$ and $\arg\max( )$, or minimal value, i.e., $\min( )$ and $\arg\min( )$, out of a set of activation values and is often dependent on how the neural post-activities are computed in the first place.
For simplicity, we focus on models that compute only a single hidden competitive layer of $d_z$ neurons. Pre-activation values are denoted as $\mathbf{h} \in \mathcal{R}^{d_{z1} \times 1}$ and post-activation values are denoted as $\mathbf{z} \in \mathcal{R}^{d_z \times 1}$. However, we note that these models could be extended to $L$ layers to create a deep competitive neural system. 

The most prominent ways of computing pre-activation values are through subtraction and distance calculation or dot products and we focus on these two in this work. The parameters for an NCL model, at minimum, include $\Theta = \{ \mathbf{M} \}$ where $\mathbf{M} \in \mathcal{R}^{d_z \times d_x}$.
If the first approach is taken, then formally the pre-activity is computed as:
\begin{align}
    \mathbf{h}[i,1] = ||\mathbf{x} - (\mathbf{M})^T[:,i]||_p \label{eqn:sub_dist}
\end{align}
where the $i$th element of the pre-activation $\mathbf{h}$ is a subtraction of the current pattern vector and the $i$th column of the transposed memory matrix, input as argument to the $p$-norm function. 
The second approach follows a form of computation typical to feed-forward artificial neural networks (ANNs):
\begin{align}
    \mathbf{h} = \mathbf{M} \cdot \mathbf{x}, \; \mbox{or, } \; \mathbf{h}[i,1] = \mathbf{M}[i,:] \cdot \mathbf{x}  \mbox{.} \label{eqn:dot_prod}
\end{align}

Once the pre-activation value $\mathbf{h}$ has been computed for all neurons in layer $1$, we may apply the best-matching post-activation function that conducts the actual required competition as shown in the simple procedure depicted in Algorithm \ref{algo:competition}. In this algorithm, we observe that the above algorithm can either compute the $K$ minimal or maximal neurons (or $K$ BMUs, stored in the list/set $w$). Furthermore, note that the above algorithm makes use of both $\max( )$/$\min( )$ and $\arg\max( )$/$\arg\min( )$ functions to obtain not only the values of winning neurons but also their indices within the activity vector $\mathbf{h}$.

\paragraph{Synaptic Weight Adjustment:}
The other fundamental function needed to fully specify an NCL system is its method for adjusting synaptic weight values given the results of the competition function (described in the last section).
Typically, the form of the update used in competitive learning is a local Hebbian or anti-Hebbian rule \cite{foldiak1990forming}. The goal of this routine, in any of the models/algorithms we examine in this paper, is to produce an update matrix $\Delta \mathbf{M}$ (of the same shape as $\mathbf{M}$) which is then used to actually alter the values within $\mathbf{M}$ as follows:
\begin{align}
    \mathbf{M} \leftarrow \mathbf{M} + \alpha \Delta \mathbf{M}
\end{align}
where $0 < \alpha < 1$ controls the magnitude of the update applied to parameters $\mathbf{M}$. Note that other rules for changing the values of $\mathbf{M}$ are possible, such as an adaptive learning rate, e.g., Adam \cite{kingma2014adam} or RMSprop \cite{tieleman2012rmsprop}.

\section{Continual Competitive Memory}
\label{sec:ccl}

In this section, we describe and formally define our proposed NCL system for task-free lifelong learning, the continual competitive memory (CCM) model. 


\subsection{Learning and Inference}
\label{sec:learning_inference}

Fundamentally, the architecture of our proposed CCM is organized into what we call ``task-memory blocks'', i.e., $\mathbf{M}_m \in \mathcal{R}^{d_{mem} \times d_x}$ is associated with a hypothetical task $m$ and $d_{mem}$ is the number of memory slots to be allocated for a given task. These blocks are all generally stored in a pool of $M$ blocks which we refer to as the ``long-term memory'' region of the system. However, at any one instant in time, one out of all these task-memory blocks is placed in what we refer to as the ``short-term memory'' region of the system. The block currently placed in short-term storage is denoted $\mathbf{M}_t$ where $t$ is known as the ``task-pointer'', an integer variable maintained by the CCM (and initialized to $t = 0$). This task pointer represents the CCM's knowledge of what current task it is operating on (out of $M$ ones currently known by the CCM).

The CCM also maintains two other important variables -- the alarm $a$, which, when exceeding integer threshold $a_\theta$, indicates a ``change of task'' and signals to the CCM to create a brand new task memory block $\mathbf{M}_{M+1}$. Once the new memory block is created, the task pointer is moved to position $M+1$.
Furthermore, a recall vector $\mathbf{r} \in \mathbb{Z}^{M \times 1}$, a vector of recall integers, one per current memory block, is maintained by the CCM. Whenever the $\max(\mathbf{r})$ exceeds a threshold $r_\theta$, the CCM is triggered to enter into ``recall'' mode and ultimately changes its task pointer $t$ to point to the long-term memory block most frequently referred to (within a window of samples).

In terms of parameters, in addition to the counter/tracking variables described above, the CCM is formally specified by the set $\Theta = \{(\Upsilon_1,\mathbf{M}_1,\mathcal{C}_1),...,(\Upsilon_M,\mathbf{M}_M,\mathcal{C}_M) \}$ where, at initialization, $M = 1$. Coupled with each actual memory block $\mathbf{M}_m$ are two lists:  $\Upsilon_m$ is a list of length $d_{mem}$ containing the per-memory learning rates (that will be decayed over after pattern is generated by the stream) and $\mathcal{C}_m$ is a list of length $d_{mem}$ where each element tracks the number of patterns readily matched to its assigned memory slot in $\mathbf{M}_m$. Note that $M$, while unbounded in this paper's implementation of the CCM, will only grow if the CCM detects a shift in the data stream's distribution (which would be a strong enough shift to indicate the presence of a new task) and only if the CCM cannot recall seeing the data it is currently processing at all (when peeking into its long-term memory using $\mathbf{r}$).

\begin{figure*}
\begin{center}
\includegraphics[width=0.825\textwidth]{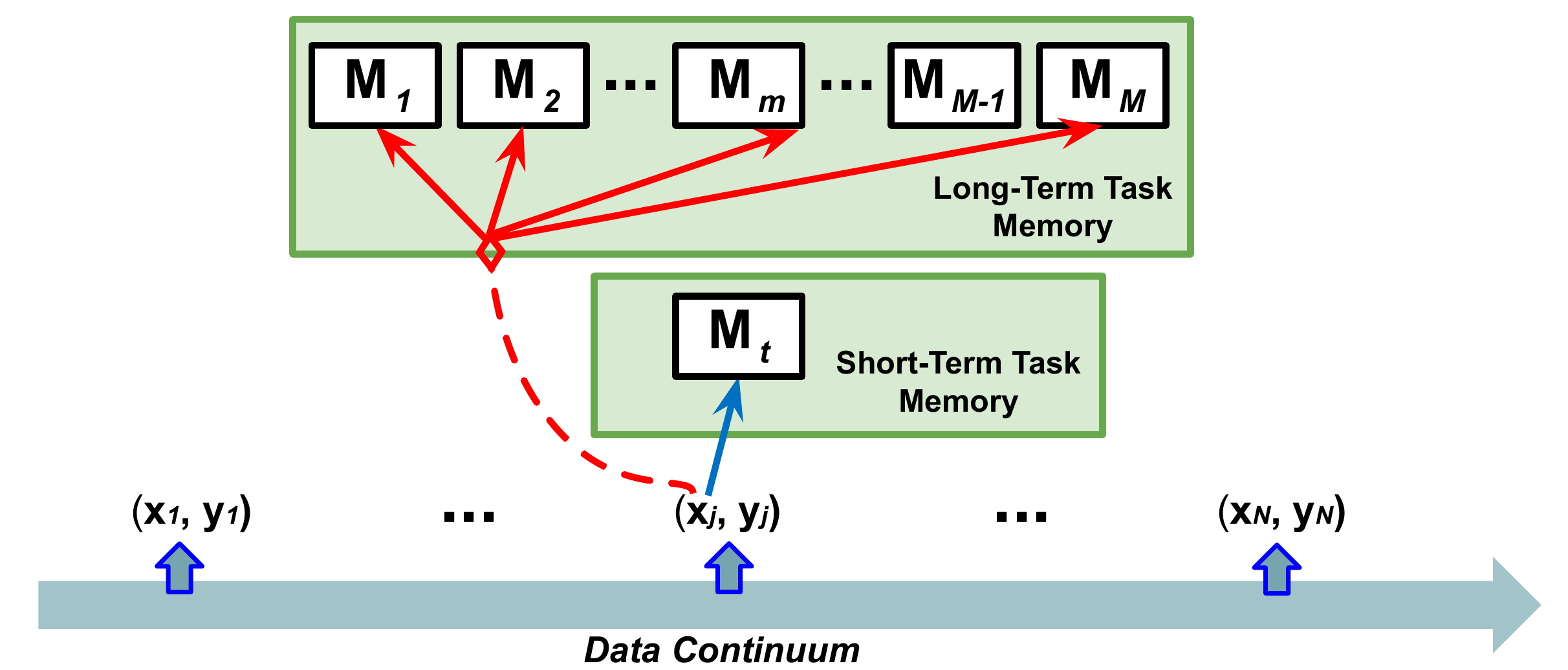}
\caption{Continual competitive memory (CCM) shown processing the $j$th pattern from a continuum.}
\label{fig:ccm_process}
\end{center}
\vspace{-0.5cm}
\end{figure*}

From a mechanics standpoint, when presented with an input pattern $\mathbf{x}$ (such as the $j$th pattern $\mathbf{x}_j$ from a stream), CCM first computes the dot product between the normalized of $\mathbf{x}$ ($\mathbf{\bar{x}}$) with each memory unit in the short-term task memory block $\mathbf{M}_t$. If the dot product of any of these potential matches exceeds $\rho$, then the CCM simply updates the matched prototype with the highest dot product/similarity score (and updates the count for that unit in $\mathcal{C}_m$ and decays its learning rate in $\Upsilon_m$). 
If none of these dot products exceed $\rho$, CCM then increments the alarm $a$ by one and then immediately checks if the current input matches any task memory units in its long-term memory. Note that if $\mathbf{x}$ matches a slot $q$ in the long term memory, exceeding the recall vigilance meta-parameter $\rho_r$\footnote{Higher values for $\rho$ and $\rho_r$, i.e., $0.8$, ensures the CCM is ``sure'' that an input belongs to task memory $\mathbf{M}_m$.}, then $\mathbf{r}[q,1]$ is incremented by one. If $\max(\mathbf{r}) > r_\theta$, then the CCM will enter into recall mode and retrieve the task memory block $q = \arg\max(\mathbf{r})$ that has received more than $r_\theta$ consecutive hits, changing its task pointer $t = q$.
If $a > a_\theta$, however, then a fresh, empty memory block $\mathbf{M}_{M+1}$ will be created and the task pointer becomes $t$ to $M+1$.

The full process (as well as additional implementation details) of the CCM's learning and inference process is presented in Algorithm \ref{algo:ccm}. Graphically, in Figure \ref{fig:ccm_process}, we depict the CCM (at a high-level) processing the $j$th pattern vector from the data stream/continuum.
One might observe that the CCM builds upon useful concepts originally proposed in adaptive resonance theory (ART) \cite{grossberg1987competitive,grossberg2013adaptive,he2004modified}\footnote{ART was originally proposed for  non-stationary data streams in mind.}, particularly its use of a ``vigilance'' meta-parameter to drive neuronal unit competition entering during the BMU search phase. In that sense, one could consider the CCM to have generalized ART to the task-free online continual learning setting with the critical distinction that the CCM focuses on different functionality with respect to sequential learning. Furthermore, CCM is structured in terms of a short and long-term memory parameter organization and utilizes a memory update rule that is rather different than the one used in ART. As we will see in our experiments later, while ART is in of itself powerful, the CCM's design is to self-organize its memory with respect to short and long-term regions proves to be important when learning from longer, high-dimensional task streams.

\begin{algorithm}[!t]
\caption{Inference and learning for the continual competitive memory (CCM) model.}
\label{algo:ccm}
\begin{algorithmic}
   \State {\bfseries Input:} sample $\mathbf{x}$,
   $t = 1$, $\Theta = \{(\Upsilon_1, \mathbf{M}_1, \mathcal{C}_1 \}$, $a = 0$, $\mathbf{r} =  \mathbf{0}$
   \State {\bfseries Constants:} $a_\theta > 0$, $r_\theta > 0$, $C_\theta > 0$, $0 < \rho < 1$, $0 < \rho_r < 1$, $0 < \epsilon \ll 1$, $0 \ll \gamma_\upsilon < 1$
   \LineComment Adjust synapses given pattern vector
   \Function{Update}{$\mathbf{x}$, $\rho$, $\epsilon$, $ \gamma_\upsilon$, $\Theta$, $t$, $a$, $\mathbf{r}$, $a_\theta$, $r_\theta$, $C_\theta$ }
        \State $\Upsilon_t, \mathbf{M}_t, \mathcal{C}_t \leftarrow \Theta[t]$
        \State $\mathbf{\bar{x}} = \mathbf{x}/(||\mathbf{x}|| + \epsilon)$, \; $\mathbf{h}_t = \mathbf{M}_t \cdot \mathbf{\bar{x}}$
        \State $h_i = -1$, $i = -1$, $C_i = \infty$, $k = 0$
        \While{$h_i \leq \rho$ $\land$ $C_i > C_\theta$ $\land$ $k < r(\mathbf{h}_t)$} \Comment Enter search mode
        \State $i = \Call{FindBMU}{\mathbf{h}_t,K=1,\text{``max''}}$, \; $h_i = \mathbf{h}_t[i,1]$, \; $C_i = \mathcal{C}_t[i]$, 
        \State $\mathbf{h}_t[i,1] = -1$, \; $k \leftarrow k + 1$
        \EndWhile
        \If{$h_i > \rho$ $\lor$ $C_i \leq C_\theta$}     \LineComment Update $i$th slot in short-term task memory block $\mathbf{M}_t$
            \State $\upsilon_i = \Upsilon_t[i]$, \; $\mathbf{M}_t[i,:] = \frac{ \mathbf{M}_t[i,:] + \upsilon_i \big( (\mathbf{\bar{x}})^T - \mathbf{M}_t[i,:] \big) }{||\mathbf{M}_t[i,:] + \upsilon_i \big( (\mathbf{\bar{x}})^T - \mathbf{M}_t[i,:] \big)||_2 + \epsilon}$, \; $\Upsilon_t[i] = \gamma_\upsilon \upsilon_i $, \; $\mathcal{C}_t[i] \leftarrow \mathcal{C}_t[i] + 1$
        \Else 
            \LineComment Check for recall in long-term memory, otherwise create new task memory block
            \State $a \leftarrow a + 1$, \; $q = -1$
            \While{$m < |\Theta|$ $\land$ $h_i \leq \rho_r$}
                \If{$j \neq t$}
                \State $\Upsilon_m, \mathbf{M}_m, \mathcal{C}_m \leftarrow \Theta[m]$, \; $\mathbf{h}_m = \mathbf{M}_m \cdot \mathbf{\bar{x}}$
                \State $i = \Call{FindBMU}{\mathbf{h}_m,K=1,\text{``max''}}$, \; $h_m = \mathbf{h}_m[i,0]$, \; $q = m$
                \EndIf
                \State $m \leftarrow m + 1$
            \EndWhile
            \If{$h_m > \rho_r$}
                \State $\mathbf{r}[m,1] \leftarrow \mathbf{r}[m,1] + 1$
            \EndIf
            \If{$\max(\mathbf{r}) > r_\theta$} \Comment Point back to recalled memory block
                \State $t \leftarrow q$, \; $a \leftarrow 0$, \; $\mathbf{r} \leftarrow \mathbf{0}$
            \ElsIf{$a > a_\theta$} \Comment Create \& point to new block $\mathbf{M}_{|\Theta|+1}$
                \State $\Upsilon_{|\Theta|+1},\mathbf{M}_{|\Theta|+1},\mathcal{C}_{|\Theta|+1} \leftarrow \Call{InitParams}{ }$ 
                \State $\Theta \leftarrow \Theta \cup \{ (\Upsilon_{|\Theta|+1}, \mathbf{M}_{|\Theta|+1}, \mathcal{C}_{|\Theta|+1}) \}$, \; $t \leftarrow |\Theta| + 1$, \; $a \leftarrow 0$, \; $\mathbf{r} \leftarrow \mathbf{0}$ 
            \EndIf
        \EndIf 
    \EndFunction
    \LineComment Determine which memory block pattern $\mathbf{x}$ belongs to
    \Function{GetTaskPointer}{$\mathbf{x}$, $\Theta$, $\epsilon$} 
        \State $\mathbf{M} = \emptyset$ 
        \For{$m = 1$ to $|\Theta|$}  \Comment Concatenate all memory blocks into one single matrix
            \State $\Upsilon_m, \mathbf{M}_m, \mathcal{C}_m \leftarrow \Theta[m]$, \; $d_{mem} = r(\mathbf{M}_m)$
            \State $(m > 1 \rightarrow \mathbf{M} = \big[ \mathbf{M},\mathbf{M}_m \big]_r) \land (m \leq 1 \rightarrow \mathbf{M} = \mathbf{M}_m)$
        \EndFor
        \State $\mathbf{\bar{x}} = \mathbf{x}/(||\mathbf{x}|| + \epsilon)$, \; $\mathbf{h} = \mathbf{M} \cdot \mathbf{\bar{x}}$
        \State $i = \Call{FindBMU}{\mathbf{h},K=1,\text{``max''}}$, \; $t_i = \lfloor i / d_{mem}) \rfloor$
        \State \textbf{Return} $t_i$  \Comment Output task memory block pointer
    \EndFunction
\end{algorithmic}
\end{algorithm}

\subsection{Detecting Tasks through Modality Pointers at Test-Time Inference}
\label{sec:linking_with_mlp}

One of the key uses of the CCM, at various points of the stream, is to, when presented with a pool of held-out test data patterns, infer which of its own current task memory blocks that the data pool belongs to. In Algorithm \ref{algo:ccm}, we present a simple procedure ($\Call{GetTaskPointer}{ }$) that does this for a single input pattern, leveraging the current state of the CCM's long-term memory. In essence, the CCM will compare the normalized input pattern $\mathbf{\bar{x}}$ to each possible memory slot in $\mathbf{M}$, which is a row-wise concatenation of all memory blocks in long-term storage. To compute which task memory block the resultant BMU index $i$ refers to, one may simply divide $i$ by the total number of unique memory slots in any $\mathbf{M}_m$, i.e., $d_{mem}$ and take the floor of the result.
Finally, when provided with a data pool, the CCM infers the pool's task by applying $\Call{GetTaskPointer}{  }$ to each pattern, storing the resulting integer $t_i$ for each in a list $\mathcal{P}$, and finally computes the modal value $t_M$ of this list of integers, i.e., $t_M = \mode(\mathcal{P})$.

\paragraph{Coupling the CCM to a Neural Predictor: }
Armed with the above process for inferring the modal task pointer at test-time, we may now briefly describe a powerful extension of the CCM when processing data from a stream. In short, imagine that we have access to a prediction function $f_{\Theta_f}(\mathbf{x})$. The function takes on the form of an ANN where a stack of nonlinear transformations, or ${ \{f_\ell(\mathbf{z}_{\ell-1};\theta_\ell)\}^L_{\ell=1} }$, is applied to the input $\mathbf{x}$. As an example, if the network is a multilayer perceptron (MLP), each transformation $\mathbf{z}_\ell = f_\ell(\mathbf{z}_{\ell-1})$ produces an output $\mathbf{z}_\ell$ from the value $\mathbf{z}_{\ell-1}$ of the previous layer with the help of a weight matrix $ \theta_\ell = \{\mathbf{W}_{(\ell-1) \rightarrow \ell}\}$. $f_\ell$  is decomposed into two operations (biases omitted for clarity):
\begin{align}
\mathbf{z}_\ell &= \phi_\ell(\mathbf{h}_\ell),\quad \mathbf{h}_\ell = \mathbf{W}_{(\ell-1) \rightarrow \ell} \cdot \mathbf{z}_{\ell-1} \label{eqn:mlp}
\end{align}
where $\phi_\ell$ is an activation function,  ${ \mathbf{z}_\ell \in \mathcal{R}^{H_\ell} }$ is the post-activation of layer $\ell$ while ${ \mathbf{h}_\ell \in \mathcal{R}^{H_\ell} }$ is the pre-activation vector of layer $\ell$ ($H_\ell$ is the number of neurons in layer $\ell$). Note that $\mathbf{z}^0 = \mathbf{x}$. 

To leverage the CCM's task pointer and couple it to the MLP predictor to prevent catastrophic interfernce when updating MLP parameters via backprop, we take inspiration from the biologically-inspired complementary neural system proposed in \cite{ororbia2019lifelong}. Specifically, we introduce one final set of parameters for each layer of the MLP (except the bottom-most and top-most layers), i.e., a binary matrix $\mathbf{Q}^\ell \in \{0,1\}^{|\Theta| \times H_\ell}$ where each row contains a unique set of $M_H$ ones ($M_H \ll H_\ell$). The CCM's task pointer, either $t$ during training within the stream or $t_M$ during test-time inference, is encoded via 
$\mathbf{t} = \Call{Enc}{t,|\Theta|}$ 
which is then multiplied with each $\mathbf{Q}^\ell$ to obtain a gating vector $\mathbf{g}^\ell$ for each hidden layer of the MLP. The gate vector is formally calculated as $\mathbf{g}^\ell = \mathbf{Q}^\ell \cdot \mathbf{t}$ which is multiplied element-wise with each layer of the MLP as follows: $\mathbf{z}^\ell \leftarrow \mathbf{z}^\ell \otimes \mathbf{g}^\ell$. 

The above coupling of the CCM with the MLP (we refer to this as CCM-MLP in our experiments), allows the CCM to drive portions or sub-networks of the MLP predictor when processing data and facilitate stronger label predictive generalization. Furthermore, this demonstrates that the CMM is compatible with popular ANNs like the MLP.

\section{Neural Competitive Learning Baselines}
\label{sec:ncl_baselines}

In this section, we describe several important NCL baseline models that our CCM will be compared against, including our adaptation of adaptive resonance theory (ART).

\subsection{The Incremental Winner-Take-All (iWTA) Model}
\label{sec:wta}

The simplest competitive learning model is the winnter-take-all (WTA) system. In this model, it is typical to combine the subtractive pre-activation rule (Equation \ref{eqn:sub_dist}) with a distance function and Algorithm \ref{algo:competition}.
The resulting model, with parameter matrix $\mathbf{M} \in \mathcal{R}^{d_z \times d_x}$ and generalized for finding the $K$ BMU given input $\mathbf{x}$, is:
\begin{align}
    \mathbf{h}[i,1] &= ||\mathbf{x} - (\mathbf{M})^T[:,i]||_p, \; \forall i = 0,1,..,\text{r}(\mathbf{M}) \\
    w &= \Call{FindBMU}{\mathbf{h}, K, \text{``min''}}, \;  \mbox{and}, \; \mathbf{z} = \mbox{Enc}(w, d_z)
\end{align}
where $K = 1$ recovers the classical unsupervised vector quantization model \cite{gray1984vector}. To update the synaptic weights of the WTA model, we employ the following Hebbian rule:
\begin{align}
    \Delta \mathbf{M}[i,:] =  \mathbf{z}[i,1] \Big( (\mathbf{x})^T - \mathbf{M}[i,:]\Big)  \mbox{.}
\end{align}
If we observe that each subtraction vector in the rule above is simply being multiplied by the $i$th value in the binary column vector $\mathbf{z}$, we can write a vectorized form of the rule as follows:
\begin{align}
    \Delta \mathbf{M} =  \big( \text{BR}_{d_z}\big( (\mathbf{x})^T \big) - \mathbf{M} \big) \otimes \text{BC}_{d_x}\big( \mathbf{z} \big) \mbox{.} 
\end{align}
Since this NCL model will adapt to patterns online, we shall further refer to it as incremental WTA (iWTA).

\subsection{The Incremental Gaussian Mixture Model (iGMM)}
\label{sec:igmm}

The iWTA model presented above can be generalized further by adapting aspects of the venerable mixture of Gaussians model, otherwise known as the Gaussian mixture model (GMM), often used in density estimation and clustering \cite{pinto2015fast,mclachlan2019finite}.
The first adaptation of the GMM to the NCL setup, which we call the incremental GMM (iGMM), assumes isotropic Gaussian components with a fixed scalar variance $\gamma = 1/\sigma^2$ (a hyper-parameter decided by the user). This yields a set of Gaussian kernels with which the input pattern $\mathbf{x}$ is compared against:
\begin{align}
   \mathbf{h}[i,1] &= ||\mathbf{x} - (\mathbf{M})^T[:,i]||_p, \; \forall i = 0,1,..,\text{r}(\mathbf{M}) \\
    \mathbf{z} &= \frac{\exp(-\gamma \mathbf{h}/T)}{\sum_j -\gamma \exp(\mathbf{h}/T)[j,1]} \quad \text{(Calculate posterior probabilities over units)}
\end{align}
where we have introduced the temperature meta-parameter $T$ (we set $T = 0.0285$ in this study).
Notice that the above formulation produces a probabilistic weight per unit/prototype rather than a hard quantization like in WTA.

The update for the iGMM also follows a simple subtractive form:
\begin{align}
    \Delta \mathbf{M} = \gamma \Big( \big( \text{BR}_{d_z}\big( (\mathbf{x})^T \big) - \mathbf{M} \big) \otimes \text{BC}_{d_x}\big( \mathbf{z} \big) \Big) \mbox{.}
\end{align}
but the prototype comparisons are weighted by the posterior probabilities over prototypes as well as the shared variance. This means that all units will be moved to some degree by the presence of an input $\mathbf{x}$.

\subsection{Adaptive Resonance Theory (ART)}
\label{sec:art}

In this section, we present the fast variant of adaptive resonance theory (ART)  \cite{grossberg1987competitive} designed to tackle real-valued/continous inputs, known as ART 2A. Specifically, we implement and explore the ART-C 2A model \cite{he2004modified}.

\begin{algorithm}[!t]
\caption{The inference and learning processes for ART-C 2A.}
\label{algo:artc2a}
\begin{algorithmic}
   \State {\bfseries Input:} sample $\mathbf{x}$, $\mathbf{M} = \emptyset$, $\Upsilon = \emptyset$, $0 < \rho < 1$
   \State {\bfseries Constants:} $0 < \upsilon_0 < 1$, $0 < \epsilon \ll 1$, $C_\theta > 0$, $0 \ll \gamma_\upsilon < 1$
   \Function{Update}{$\mathbf{x}$, $\rho$, $\epsilon$,  $\mathbf{M}$, $C_\theta$} 
        \State $\mathbf{\bar{x}} = \mathbf{x}/(||\mathbf{x}|| + \epsilon)$, \; $\mathbf{h} = \mathbf{M} \cdot \mathbf{\bar{x}}$
        \If{$\mathbf{M} \neq \emptyset$} 
            \State $h_i = -1$, $i = -1$
            \While{$h_i \leq \rho$ $\land$ $k < r(\mathbf{h})$} \Comment Enter search mode
            \State $i = \Call{FindBMU}{\mathbf{h},K=1,\text{``max''}}$, \; $h_i = \mathbf{h}[i,1]$  \Comment Find current BMU
            \State $\mathbf{h}[i,1] = -1$, \; $k \leftarrow k + 1$
            \EndWhile
            \If{$h_i > \rho$} \Comment Enter resonance mode
                \State $\upsilon_i = \Upsilon[i]$, \; $\mathbf{M}[i,:] = \frac{\upsilon_i (\mathbf{\bar{x}})^T + (1 - \upsilon_i) \mathbf{M}[i,:]}{||\upsilon_i (\mathbf{\bar{x}})^T + (1 - \upsilon_i) \mathbf{M}^1[i,:]||_2}$, \; $\Upsilon[i] = \gamma_\upsilon \upsilon_i$  \Comment Update weight vector $i$
            \Else
                \State $\mathbf{M} = \big[ \mathbf{M},(\mathbf{\bar{x}})^T \big]_r$ \Comment Conduct mismatch reset
            \EndIf 
            \If{$r(\mathbf{M}) > C_\theta$}
                \State Apply constraint reset satisfaction according to \cite{he2004modified}, yielding corrected $\mathbf{M}$
                \State Compute new vigilance $\rho$ according to \cite{he2004modified}
            \EndIf
        \Else 
            \State $\mathbf{M} = (\mathbf{\bar{x}})^T$, $\Upsilon = \{\upsilon_0\}$  \Comment Set very first weight vector to be input pattern
        \EndIf
    \EndFunction
\end{algorithmic}
\end{algorithm}

In standard ART, a vigilance meta-parameter $rho$ is used to detect novelty -- if a match computed between an input pattern and a memory unit exceeds $rho$ then the memory is adjusted (using a moving-average like rule), otherwise, a new category/memory is created (initialized to the current pattern).
In Algorithm \ref{algo:artc2a}, our implementation of the ART-C 2A process as an NCL model also notably includes an extension known as constraint reset, meant to prevent an unbounded growth of memory units \cite{he2004modified}.
ART-C 2A involves a bit more than just calculation of the BMU post-activation before updating weights -- it, as does CCM, conducts a search through its activities to find a BMU that satisfies a ``vigilance'' threshold check. If none of the currently available activities pass the vigilance check, then a new category/weight vector is instantiated. The constraint reset mechanism, again, ensures that this NCL model never has more than $C_\theta$ units. To satisfy this bound on units, once the number of units exceed $C_\theta$, a merging operation is applied that replaces the two nearest neighbor memories (in terms of dot-product similarity) and replaces them both with a single vector containing their average. Furthermore, the vigilance parameter is dynamically re-computed upon each input presentation  \cite{he2004modified} (based on the maximum vigilance computed given the current input $\mathbf{x}$). 
Note that we modified the underlying ART-C 2A process further by incorporating a per-unit learning rate that was decayed as a function of the number of input samples similar to that used in our CCM (we found this improved generalization a bit when learning online).

We comment that the literature related to ART is vast and a plethora of model variants exist. Given that it was our interest to only examine one of the simplest/fastest variants of ART, i.e., ARTC-2A, when making comparisons, the reader should read the results reported for our ART model noting that potentially one of the more complex/intricate ART systems (such as those based on differential equations \cite{grossberg1987competitive}) might yield different and different results.

\subsection{Modal Classification}
\label{sec:classify}
All of the NCL models presented above, including our CCM, can further be extended to offer some basic discriminative functionality. While some models like ART have special generalizations such as ARTMAP \cite{carpenter1991artmap}, we will focus, in this study, on generalizing all of the unsupervised memory systems above in the same way.
Specifically, an NCL model can be made to conduct classification of its input through the following two alterations:
\begin{itemize}
    \item During learning/synaptic adjustment, the NCL model must maintain and update a single frequency vector $\mathbf{y}^c_i \in \mathbb{Z}^{d_y \times 1}$ for each memory unit. meaning there are $d_z$ (or $d_{mem} \times d_z$ for the CCM) total frequency vectors yielding an extra tensor parameter $\mathbf{Y}^c \in \mathbb{Z}^{d_z \times d_y \times 1}$. This means that, upon presentation of $(\mathbf{x}_j, \mathbf{y}_j)$, the model computes $\mathbf{h}$ and $i = \Call{FindBMU}{\mathbf{h},K,\text{winner\_type}}$ and then updates $i$th slice of $\mathbf{Y}^c$ through simple addition, i.e., $\mathbf{Y}^c[i,:,:] = \mathbf{Y}^c[i,:,:] + \mathbf{y}_j $ (since $\mathbf{y}_j$ is one-hot encoded, it already contains an increment of one at the correct class index for sample $j$). This means $\mathbf{y}_j$ is bound to its relevant unit.
    \item During (test-time) inference, upon the presentation of $\mathbf{x}_j$, the NCL model classifies the pattern by first calculating pre-activity $\mathbf{h}$ as well as the winning neuron $i = \Call{FindBMU}{\mathbf{h},K,\text{winner\_type}}$ and then retrieving the $i$th slice of $\mathbf{Y}^c$. The final label assignment is made via the following rule:  $y_j = \arg\max_y \mathbf{Y}^c[i,:,:]$ where $y_j$ is an integer representing the NCL model's predicted class for $\mathbf{x}_j$. 
 \end{itemize}
Note that the above two modifications to our NCL models means that we are conducting classification by estimating the target label as the mode of the known training sample labels bound over time to the relevant memory $\mathbf{h}[:,i]$ that responds to $\mathbf{x}_j$, i.e., assigning the majority vote label. This choice of mechanism allows us to separate out the classification process from the pattern matching process of the memory system, providing us with a general family of models that can handle the case of data streams that either have no labels at all (unsupervised) or present some samples with labels and others without (semi-supervised).

We note that there are other mechanisms that could be used instead to conduct classification (and potentially yield better discriminative performance), including modifying the learning process of each NCL to work similar to learning vector quantization \cite{kohonen1995learning} or to learn a classifier jointly with the memory system (as in the case of ARTMAP \cite{carpenter1991artmap}). Nonetheless, we shall leave investigation of such generalizations for future work given that our interest in this paper is to investigate memory retention without joint reliance on the label.

\section{Experiments}
\label{sec:experiments}

\subsection{Metrics}
\label{sec:metrics}
In this section, we describe the set of evaluation metrics used to compare our CCM against the competitive learning baselines. Specifically, we measure performance with respect to two properties: 
1) knowledge retention and generalization across tasks in a stream, and 
2) cluster validity through class conformity.

\paragraph{Generalization and Knowledge Retention:}
In order to quantify each model's ability to generalize and retain previous knowledge, we measure average accuracy (ACC), or mean performance across tasks, and backwards transfer (BWT), two key metrics often used to quantify how well a continual learning model retains knowledge. 

To use these metrics, we must first compose a task matrix $R$ (as in \cite{lopez2017gradient}), which is an $T \times T$ matrix of task accuracy scores (normalized to $[0,1]$). With respect to $R$, the two key metrics we measure are defined as:
\begin{align}
\mbox{\normalsize ACC} = \frac{1}{T}
\sum_{i=1}^T R_{T,i},\quad
\mbox{\normalsize BWT} = \frac{1}{T-1}
\sum_{i=1}^{T-1} R_{T,i} - R_{i,i}  \mbox{.} \label{eq:acc_and_bwt}
\end{align}
BWT measures the influence that learning a task $i$ has on the performance of task $k < i$. A positive BWT indicates that a learning task $i$ increases performance on a preceding task $k$. As such, a higher BWT is better and a strongly negative BWT means there is stronger (more catastrophic) forgetting.

\paragraph{Class Conformity:}
We next measure how well the clusters acquired by each model conform to the target class distribution associated with each benchmark. Since our NCL models are applied to supervised learning streams, we are interested in examining explicit clustering metrics that determine how similar each model's clustering it to the input data's (empirical) ground truth label distribution. In this paper, we record three such explicit metrics (bounded in the range of $[0,1]$): adjusted random index (ARI) \cite{hubert1985comparing}, the Fowlkes-Mallows index (FMI) \cite{fowlkes1983method}, and the V-measure (V-M) \cite{rosenberg2007v} (we refer the reader to the references for implementation details of each metric). 

For all three of the above conformity metrics, a higher score closer to $1.0$ is desired. 
Values closer to zero generally indicate that a model's clustering and the ground truth labels are largely independent while values closer to one indicate significant agreement. Furthermore, we selected these three metrics given that all of them make no assumption as to what kind of cluster structure that any particular algorithm should yield. Notably, the V-measure was used in \cite{he2004modified} (which proposed ARTC-2A) and is derived from conditional entropy analysis, offering a measurement that accounts for two desirable properties one would want in a model's resultant clustering: homogeneity, where each cluster contains solely members of a single class/label, and completeness (class entropy), where all members of a particular class are assigned to the same cluster by a given model.




\begin{table}[t]
\begin{center}
\begin{tabular}{l | c | c | c  } 
 \textbf{MNIST} & FMI & ARI & V-M \\
 \hline\hline
 iWTA & $0.7464 \pm 0.004$ & $0.7179 \pm 0.0044$ & $0.8234 \pm 0.0023$  \\  
 iGMM & $0.7526 \pm 0.0292$ & $0.7249 \pm 0.0323$ & $0.8256 \pm 0.0175$  \\ 
 ART-C 2A & $0.6635 \pm 0.0245$ & $0.6237 \pm 0.0285$ & $0.7817 \pm 0.0129$  \\ 
 CCM (Ours) & $\mathbf{0.8990 \pm 0.0003}$ & $\mathbf{0.8877 \pm 0.0004}$ & $\mathbf{0.9132 \pm 0.0002}$  \\
 \hline 
 \textbf{NotMNIST} & FMI & ARI & V-M \\
 \hline
 iWTA & $0.6896 \pm 0.0156$ &$0.6551 \pm 0.0174$ & $0.7887 \pm 0.0080$ \\  
 iGMM & $0.7014 \pm 0.0213$ & $0.6680 \pm 0.0236$ & $0.7967 \pm 0.0122$ \\ 
 ART-C 2A & $0.6495 \pm 0.0205$ & $0.6096 \pm 0.0221$ & $0.7716 \pm 0.0115$ \\ 
 CCM (Ours) &  $\mathbf{0.7319 \pm 0.0114}$ & $\mathbf{0.7021 \pm 0.0126}$ & $\mathbf{0.8111 \pm 0.0060}$ \\[1ex] 
\end{tabular}
\caption{Class conformity measurements - mean \& standard deviation reported for $10$ trials.}
\label{tab:conformity_results}
\begin{tabular}{l | c | c } 
  & \multicolumn{2}{c}{\begin{tabular}[x]{@{}c@{}}\textbf{MNIST}\\\end{tabular}}  \\ [0.5ex] 
 & ACC & BWT  \\
 \hline\hline
 ICarl \cite{ororbia2019lifelong} & $0.9400 \pm 0.4100$ & $-0.1000 \pm 0.0040$  \\
 Mnemonics \cite{ororbia2019lifelong} & $0.9600 \pm 0.3200$ & $-0.9910 \pm 0.0050$ \\
 MLP & $0.6132 \pm 0.0133$ & $-0.4671 \pm 0.0165$  \\ 
 \hline
 iWTA & $0.8285 \pm 0.0022$ & $-0.0824 \pm 0.0122$  \\   
 iGMM & $0.8329 \pm 0.0145$ & $-0.0746 \pm 0.0231$  \\  
 ART-C 2A & $0.7220 \pm 0.0216$ & $-0.1829 \pm 0.0164$  \\ 
 CCM & $0.9418 \pm 0.0011$ & $-0.0211 \pm 0.0001$ \\
 CCM-MLP & $\mathbf{0.9853 \pm 0.0002}$ & $\mathbf{-0.0006 \pm 0.0001}$ \\
 \hline 
 & \multicolumn{2}{c}{\begin{tabular}[x]{@{}c@{}}\textbf{NotMNIST}\\\end{tabular}} \\
 \hline
 ICarl \cite{ororbia2019lifelong} & $0.8870 \pm 0.1020$ & $-0.1090 \pm 0.0070$  \\
 Mnemonics \cite{ororbia2019lifelong} & $0.9500 \pm 0.0710$ & $-0.0110 \pm 0.0070$  \\
 MLP & $0.6222 \pm 0.0243$ & $-0.4389 \pm 0.0336$  \\ 
 \hline
 iWTA & $0.8386 \pm 0.0068$ & $-0.0468 \pm 0.0109$ \\   
 iGMM & $0.8437 \pm 0.0128$ & $-0.0350 \pm 0.0048$ \\  
 ART-C 2A & $ 0.7852 \pm 0.0164$ & $-0.0755 \pm 0.0028$ \\ 
 CCM & $0.8784 \pm 0.0110$ & $-0.0074 \pm 0.0081$\\
 CCM-MLP & $\mathbf{0.9553 \pm 0.0006}$ & $\mathbf{-0.0024 \pm 0.0019}$\\[1ex] 
\end{tabular}
\caption{Knowledge retention metrics. Mean \& standard deviation reported over $10$ trials.}
\label{tab:lifelong_results}
\end{center}
\vspace{-0.5cm}
\end{table}

\subsection{Data Benchmarks for Online Lifelong Learning}
\label{sec:online_lml}


\paragraph{Split Image Benchmarks:}
\label{sec:split_benchmarks}

The MNIST dataset\footnote{Available at the URL:  http://yann.lecun.com/exdb/mnist/.} 
contains $28\times28$ images with gray-scale pixel feature values in the range of $[0,255]$. The only preprocessing applied to this data is to normalize the pixel values to the range of $[0,1]$ by dividing them by 255.
NotMNIST\footnote{URL: http://yaroslavvb.blogspot.com/2011/09/notmnist-dataset.html} is a more difficult variation of MNIST created by replacing the digits with characters of varying fonts/glyphs (letters A-J). The preprocessing and training, validation, and testing splits were created to match the setup of MNIST with the exception that there is more data in NotMNIST, i.e., $100,000$ data points are in the training split we used for our experiments (adapted from the smaller variant in \cite{ororbia2020continual}).

To create Split MNIST and Split NotMNIST, the samples in each are organized by class. A series of five tasks ($t = \{1,2,3,4,5\}$) are created as follows:
1)  \textbf{Task} $1$ -- categorize the first two classes, e.g., $0$ and $1$ in MNIST,
2) \textbf{Task} $2$ -- categorize the next two classes, e.g., $2$ and $3$ in MNIST,
3) \textbf{Task} $3$ -- categorize the next two classes, e.g., $4$ and $5$ in MNIST,
4) \textbf{Task} $4$ -- categorize the next two classes, e.g., $6$ and $7$ in MNIST, and,
5) \textbf{Task} $5$ -- categorize the last two classes, e.g., $8$ and $9$ in MNIST.
Note that for each task above, the label would be one-hot encoded into a binary vector of size $2$, i.e., $\mathbf{y}_j \in \{0,1\}^{2 \times 1}$. Mini-batches of samples produced by each data stream were constrained to be $1$ to simulate online learning (data in each task was shuffled only once prior to simulation).

For baseline models, we not only compared to a tuned implementation of each of the NCL baseline models (described in Section \ref{sec:ncl_baselines}) but to also an MLP with two hidden layers (first layer had $110$ units and second layer had $104$ units -- these numbers were selected to ensure the MLP had the same number of total synaptic parameters as the NCL models). For iWTA and iGMM, the total number of memory units were set to $125$ (which would be the maximum number of units allowed that both ART-C 2A and the CCM could grow by the upon reaching the end of the stream -- for ART-C 2A, we set $C_\theta = 125$). For both CCM and ART-C 2A, $\epsilon = 0.00001$. For ART-C 2A, we set $\gamma_\upsilon = 1$ and $\upsilon_0 = 0.02$ for MNIST and $\gamma_\upsilon = 0.998$ and we set $\upsilon_0 = 0.1$ for NotMNIST. For the CCM, for MNIST, we set $a_\theta = 20$ and $\rho = \rho_r = 0.8$ while, for NotMNIST, we set $a_\theta = 30$ and $\rho = \rho_r = 0.755$. For both datasets, the CCM used $\gamma_\upsilon = 0.998$, $\upsilon_0 = 0.35$, $r_\theta = 30$, and $C_\theta = 60$.

\begin{figure}
\centering
\begin{subfigure}[b]{.315\linewidth}
\includegraphics[width=\linewidth, frame]{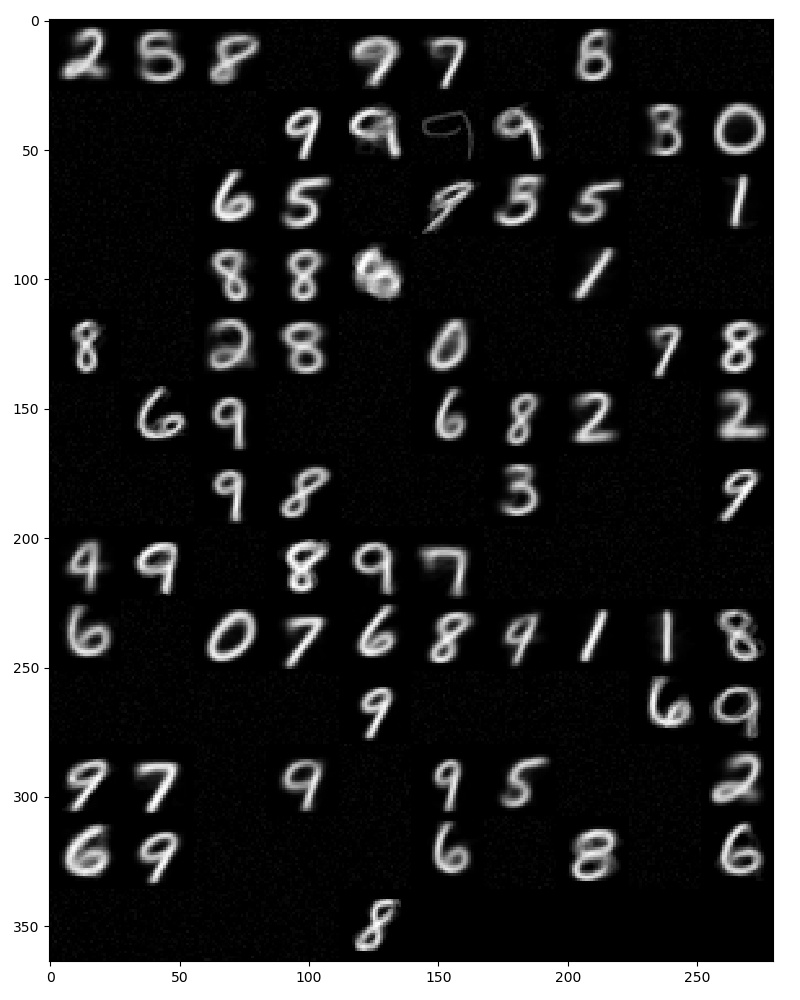}
\caption{iGMM.}\label{fig:igmm_mnist}
\end{subfigure}
\hspace{0.1cm}
\begin{subfigure}[b]{.315\linewidth}
\includegraphics[width=\linewidth, frame]{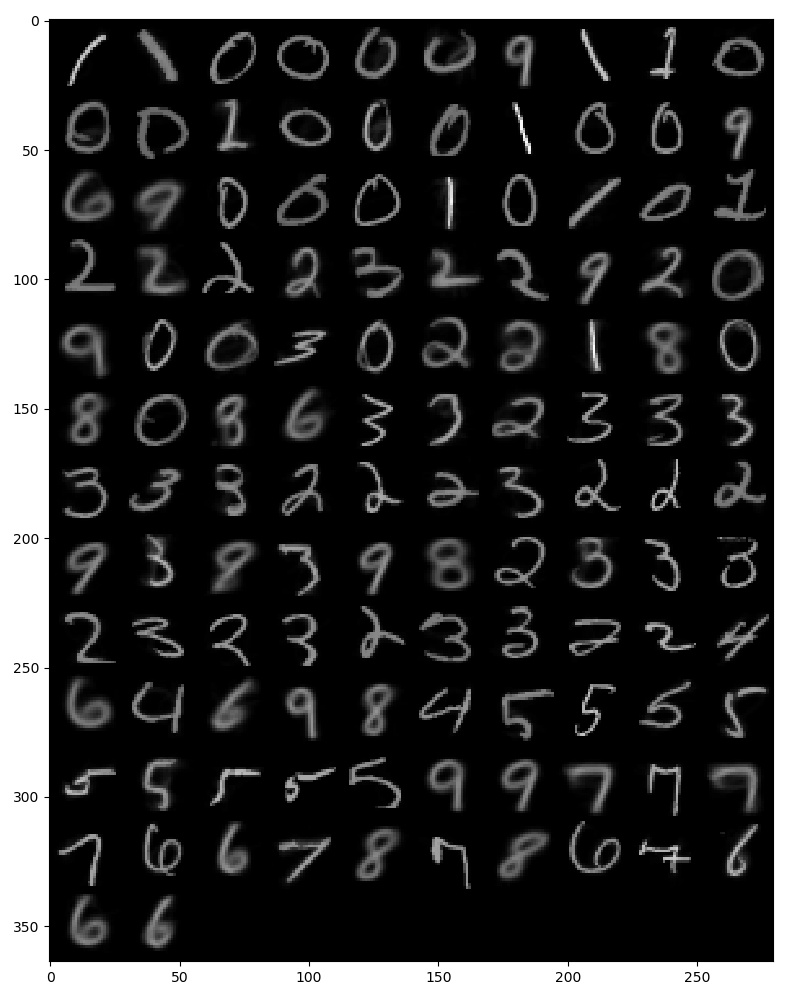}
\caption{ART-C 2A.}\label{fig:artc_mnist}
\end{subfigure}
\hspace{0.1cm}
\begin{subfigure}[b]{.315\linewidth}
\includegraphics[width=\linewidth, frame]{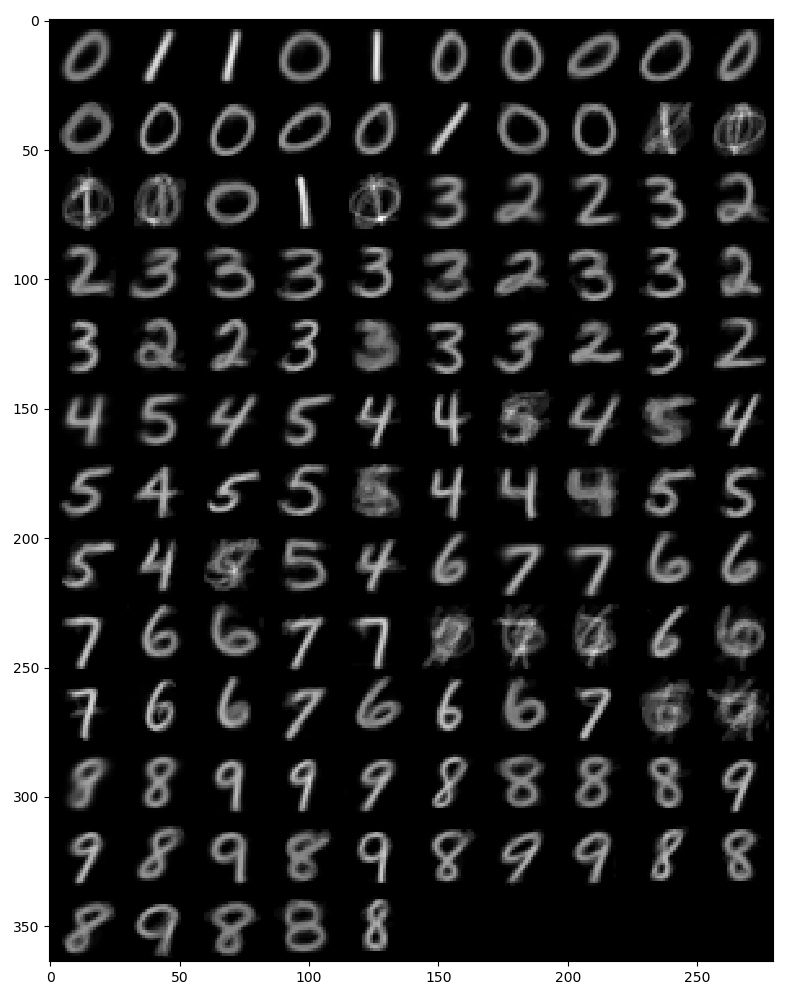}
\caption{CCM.}\label{fig:ccm_mnist}
\end{subfigure}
\begin{subfigure}[b]{.315\linewidth}
\includegraphics[width=\linewidth, frame]{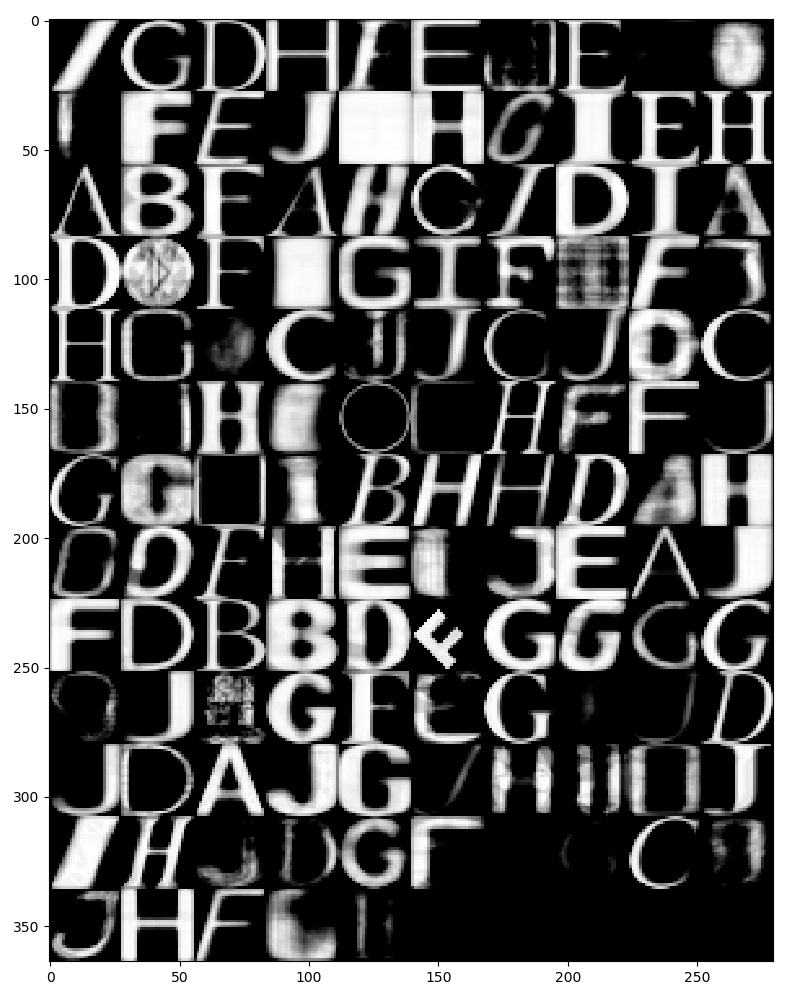}
\caption{iGMM.}\label{fig:igmm_nmnist}
\end{subfigure}
\hspace{0.1cm}
\begin{subfigure}[b]{.315\linewidth}
\includegraphics[width=\linewidth, frame]{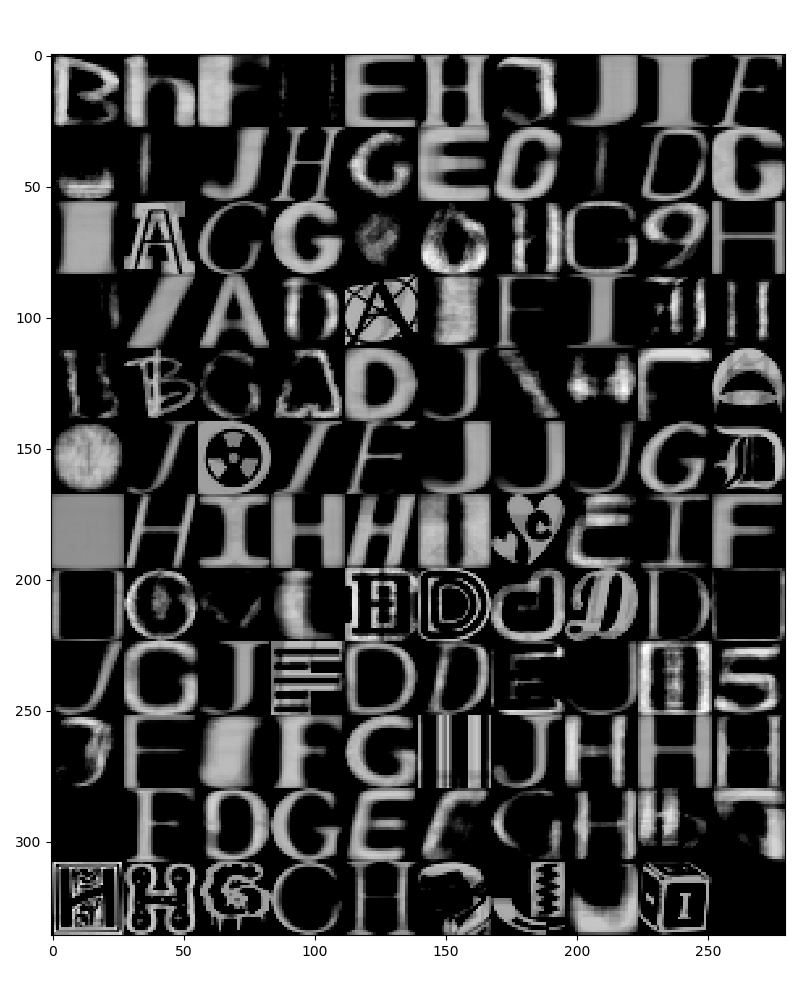}
\caption{ART-C 2A.}\label{fig:artc_nmnist}
\end{subfigure}
\hspace{0.1cm}
\begin{subfigure}[b]{.315\linewidth}
\includegraphics[width=\linewidth, frame]{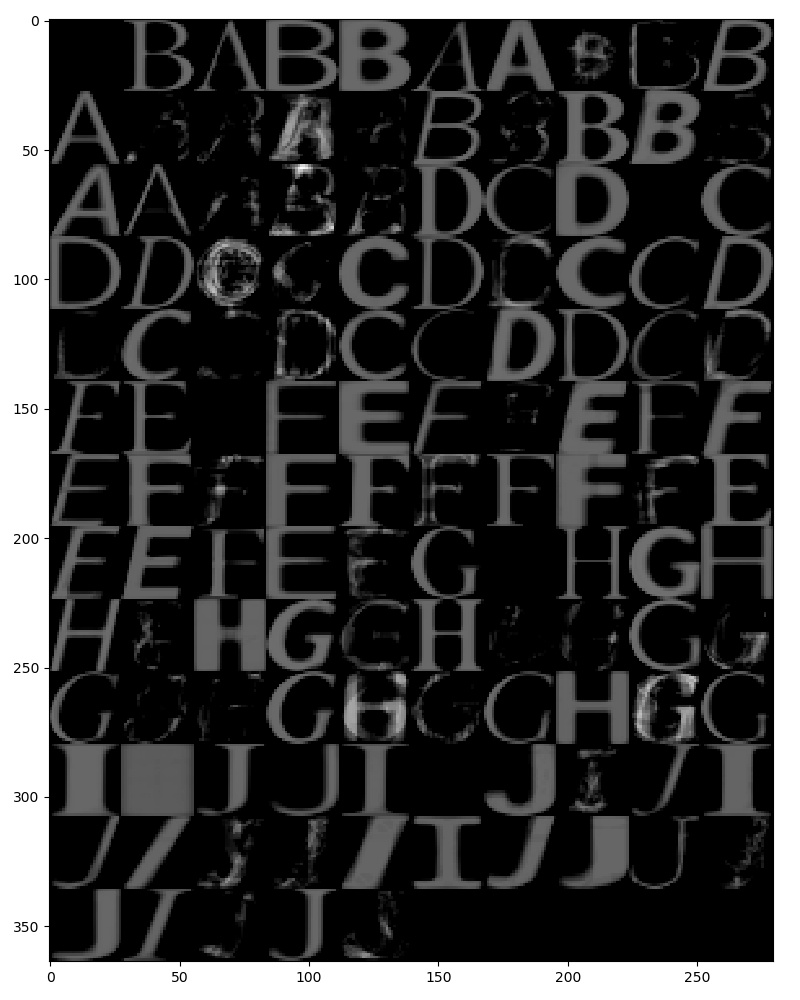}
\caption{CCM.}\label{fig:ccm_nmnist}
\end{subfigure}
\caption{Sample memories for each neural model -- top row are the memories for MNIST and the bottom row are the memories for NotMNIST. Note that ART-C 2A and CCM memories are normalized such that they have unit Euclidean norms whereas the iGMM memories are not.}
\label{fig:ncl_memories}
\vspace{-0.5cm}
\end{figure}

\subsection{Results}
\label{sec:results}

In Tables \ref{tab:conformity_results} and \ref{tab:lifelong_results}, we present our main results for the various NCL models evaluated for this study. We observe that, for all metrics (both for class conformity and knowledge retention/generalization) measured, our CCM not only outperforms the MLP baselines and other NCL models but offers strong resistance to forgetting as indicated by its high ACC and low BWT scores (for both the CCM and coupled CCM-MLP model).
Another promising result is that, with respect to the ACC and BWT, all of the NCL models improve on memory retention and generalization compared to the MLP trained with backprop. This promising result offers empirical evidence that competitive learning can serve as a means of learning semi-distributed representations that combat excessive neural cross-talk and, ultimately, catastrophic interference. Notice that at the top of Table \ref{tab:lifelong_results}, we have included two modern-day, state-of-the-art MLP baselines trained via backprop -- for our CCM using modal classification, we most nearly match the ICarl model and for the CCM-MLP integrated model, we outperform even these competitive baselines.

In Figure \ref{fig:ncl_memories}, we visualize the prototypes acquired by the iGMM, ART-C 2A, and the proposed CCM. While all of the models have appeared to acquire memory slots specialized for each of the digits (MNIST) or letters (NotMNIST), we notice that the iGMM does not fully utilize most of its capacity to represent data from each task, especially in the case of Split MNIST. We hypothesize that this might be due to the fact that, since all of the iGMM's memory slots are adjusted in the presence of sensory input (proportional to their estimated posterior probabilities), there is less diversity among the different memory units (since constantly moving all units might prevent many from specializing).
Furthermore, we observe that the CCM's memories, as expected since its design embodies the inductive bias of a dual short-term and long-term memory organization of parameter blocks, offer a more balanced pool of digits from each class per task. This is likely a driver behind its high classification performance in general.

\section{Related Work}
\label{sec:lit_review}

Artificial neural systems that conduct inference and learning based on competitive learning has a fairly long history \cite{rumelhart1985feature,desieno1988adding,wang1997competitive}. In essence, competitive learning centers around the idea that a pool of neurons/units compete for the right to activate -- leading to different units that activate for different bundles or clusters of patterns, i.e., yielding a form of neural template matching. As more data is presented to this pool of neurons, each unit in the pool will ultimately converge to the center of the cluster it has come to model. This means that each neuron will activate more strongly for sensory inputs strongly related to its cluster and more weakly for patterns related to other units/clusters. Furthermore, there exists a vast array of neural models and their learning algorithms based on competition-oriented neural computation. These models range from specialized supervised systems driven by learning vector quantization \cite{gray1984vector} to explicit topology clustering systems based on the esteemed self-organizing map (SOM) \cite{kohonen1990self}. Other models have generalized competitive learning in other directions, such as those that make use of the top $K > 1$ winning neurons as in the case of models that classify under competitive Hebbian learning \cite{white1991competitive,martinetz1993competitive}, including even compression systems based on neural gas and its variants \cite{martinetz1993neural,fritzke1995growing}. Many if not all of these models can be reformulated as NCL models. In addition, the NCL framework could be extended to account for unique features such as explicit constraints placed on the topology of competing neurons in order to include SOMs.

Adaptive resonance theory (ART) \cite{grossberg1987competitive}, from which our CCM certainly draws inspiration from, is a large and ever-growing family of powerful NCL models that favorably offer connections to real underlying neuro-biological processes \cite{grossberg2013adaptive}. While we only presented how one of the more classical (yet fast and efficient) forms of ART fall under the NCL framework in this paper, i.e., ART-C 2A, other models such as Fuzzy ART \cite{carpenter1991fuzzy} and ARTMAP \cite{carpenter1991artmap} could be similarly included, provided that sufficient modifications were made to the framework in order to account for special features unique to each variant.

In terms of investigating and generalizing competitive learning effectively to the grand challenge of task-free lifelong machine learning, far less work exists. Of the potential methods developed in prior efforts, the self-organizing incremental neural network (SOINN) \cite{furao2007enhanced,shen2008fast} represents one particular system that is also meant to operate in the face of data streams (including data that comes from tasks presented sequentially). SOINN is, in some sense, an interesting generalization of growing neural gas (GNG) \cite{fritzke1995growing}, where neural prototypes are generated incrementally based on several criterion and are, crucially, guided by a relation graph between the units that is incrementally constructed and constantly updated. 
However, while this competitive learning model is promising for processing data streams, it is difficult to scale it to large problem spaces given that an entire graph of the units/centroids must be maintained and a vast array of key meta-parameters are introduced that must be carefully tuned to the problem space at hand. Furthermore, the experiments shown to evaluate the efficacy of the SOINN, while again yielding positive results, tend to focus on rather small-scale, low-dimensional problems/datasets.

\section{Conclusions}
\label{sec:conclusion}
In this paper, we examined several powerful, biologically-plausible competitive neural systems, unifying them under a common computational framework, i.e., the neural competitive learning (NCL) framework, and applied them to the challenging scenario of online continual learning, without using explicitly provided/set task descriptors.
Furthermore, we proposed a novel neural memory based on competitive learning that we call continual competitive memory (CCM). Our results demonstrate that not only does our proposed CCM model generalize well to the task of online continual learning, offering performance competitive with modern-day continual learning baselines, but that competitive learning, in general, provides some robustness to catastrophic interference as compared to an artificial neural network trained via back-propagation of errors.
While powerful, the CCM does not come without its limitations: 
1) it rejects data samples with a fixed threshold (a scheme to dynamically adjust the CCM's thresholds would be highly desirable), and,
2) it does not offer an upper bound/constraint on the number of units it grows as it processes data from a stream. This last issue could be resolved by developing a constraint reset schema much like that employed in the adaptive resonance theory model we experimented with in this paper, however, great care would need to be taken to ensure that merging of units is done in such a way to not disrupt the long-term memory too much and risk accidental deletion of important task-specific knowledge. Finally, given that we have shown that the CCM readily integrates with other predictors to improve generalization, future work should entail investigating replacing the task selection basal ganglia model explored in \cite{ororbia2019lifelong} with our CCM, since the CCM overcomes many of the limitations faced by the original task selector, yielding a much more robust, neuro-cognitively plausible continual learning system. 

\bibliographystyle{acm}
\bibliography{ref}

\end{document}